\documentclass{isprs} % isprs class modified 23-04-2019 (Dennis Wittich)
\usepackage{subfigure}
\usepackage{setspace}
\usepackage{geometry} % added 27-02-2014 Markus Englich
\usepackage{epstopdf}
\usepackage[labelsep=period]{caption}  % added 14-04-2016 Markus Englich - Recommendation by Sebastian Brocks
\usepackage[british]{babel} 
\usepackage[hang]{footmisc}
\usepackage{amsmath}
\begin{document}

\title{Adaptive Scaling with Geometric and Visual Continuity of completed 3D objects}
\date{}

% KAO: Remove extra spacing

% KAO: Remove extra spacing
\author{
 Jelle Vermandere\textsuperscript{1}, 
 Maarten Bassier\textsuperscript{1},
 Maarten Vergauwen\textsuperscript{1}
}

% KAO: Remove extra newline
\address{
	\textsuperscript{1} KU Leuven, Department of Civil Engineering, Ghent, Belgium \\
    (jelle.vermandere, maarten.bassier, maarten.vergauwen)@kuleuven.be\\
}

% KAO: Use times symbol
\abstract{
Object completion networks typically produce static Signed Distance Fields (SDFs) that faithfully reconstruct geometry but cannot be rescaled or deformed without introducing structural distortions. This limitation restricts their use in applications requiring flexible object manipulation, such as indoor redesign, simulation, and digital content creation. We introduce a part-aware scaling framework that transforms these static completed SDFs into editable, structurally coherent objects. Starting from SDFs and Texture Fields generated by state-of-the-art completion models, our method performs automatic part segmentation, defines user-controlled scaling zones, and applies smooth interpolation of SDFs, color, and part indices to enable proportional and artifact-free deformation. We further incorporate a repetition-based strategy to handle large-scale deformations while preserving repeating geometric patterns. Experiments on Matterport3D and ShapeNet objects show that our method overcomes the inherent rigidity of completed SDFs and is visually more appealing than global and naive selective scaling, particularly for complex shapes and repetitive structures.
}


\maketitle

% ---- MAIN TEXT ----
\section{Introduction}
\label{sec:introduction}
% Problem statement: 
% The need for dynamic objects 
Dynamic and adaptable object modelling from existing indoor scenes plays a critical role in various fields, including the Architecture, Engineering, Construction, and Operations (AECO) industry, where renovations and simulation design demand interactive object representations that can be manipulated, scaled and moved around. Similarly, gaming environments increasingly require more realistic and dynamic objects for immersive experiences. \cite{vermandere_guided_2025}

% object completion
Objects captured from real-world indoor environments using current remote sensing techniques are often incomplete due to sensor limitations, self-occlusions or other objects obstructing the line of sight to the sensor. However, most applications require complete watertight models. Therefore, there is a real need for completion techniques to fill in the missing parts of these incomplete scans.

% SDF techniques
Recent advancements in deep learning based object completion and generation \cite{mittal_autosdf_2022}, \cite{vasu_hybridsdf_2022}, \cite{hao_dualsdf_2020}, \cite{vermandere_geometry_2025} have resulted in a rise of the use of Signed Distance Functions (SDFs) as their final outputs. SDFs implicitly define an object's boundary by mapping the distance to the nearest surface from a given point in space \cite{mittal_autosdf_2022}. This representation can also be extended to encode texture information via texture fields \cite{oechsle_texture_2019}, enabling a one-to-one connection between surface distance and colour at any point in space.

% static objects
The resulting completed objects are typically static and lack any part awareness. Scaling these objects uniformly introduces undesirable distortions. This limitation creates challenges in adapting captured objects for diverse use cases. A key unmet need is the ability to scale objects selectively, ensuring that only specific parts are altered while retaining proportionality and visual coherence.

% procedural mesh generation
Current  state-of-the-art part-aware scaling methods are predominantly employed in the gaming industry for procedural large-scale object generation. These techniques often rely on a pre-existing user-defined list of modular parts \cite{li_proc-gs_2024}, or use a user-defined area without part-awareness \cite{deftly_27_2021}, making generalization more difficult.

In this paper, we propose a novel approach to create dynamic and scalable objects from the SDF's created by state-of-the-art object completion networks by:

\begin{itemize}
    \item Dividing the completed object into distinct parts, enabling selective scaling.
    \item Creating a framework to define specific areas of the objects that need to be scaled.
    \item Interpolating the distances and colours in the defined zones to create a seamless object. 
    \item Introducing repeating patterns for large deformations.
\end{itemize}

This approach ensures that objects can be scaled proportionally while preserving structural coherence and visual quality, addressing a critical gap in current object scaling methodologies.

% The structure
The remainder of this work is structured as follows. The background and related work is presented in Section \ref{sec:background}. Following is the explanation of the proposed method in Section \ref{sec:methodology}. In Section \ref{sec:experiments}, an overview of the used datasets and their results is presented. Finally, the conclusions are presented in Section \ref{sec:conclusion}

%=====================BACKGROUND=====================%
\section{Background and related work}
\label{sec:background}

Object scaling and deformation have been explored across various domains, including image processing, mesh deformation, procedural generation, and SDFs. In this section, we review prior work in these areas, highlighting the limitations addressed by our method.

\subsection{Object Completion}
Recent advancements in object geometry completion have shifted towards completing partial SDFs. Because they can be easily discretised into a voxel grid, they are the ideal input for machine-learning-based models like AUTO SDF\cite{mittal_autosdf_2022}, which trained a model on sub-selections of the voxel grid of complete objects. The model can then predict the missing sub-selections to complete the missing parts of the object. XCUBE \cite{ren_xcube_2023} Improves upon this method by introducing a hierarchical voxel octree representation allowing for a coarse to fine completion network which results in a much higher output resolution.
To improve the completion results for objects with more realistic occlusions, more recent works \cite{vermandere_geometry_2025} have focussed on adding more refinement to the voxel-based inputs.

% add more references
Texture completion models have tried to adapt the same function-based representation with the introduction of Texture fields \cite{oechsle_texture_2019} by encoding the texture in 3D space instead of on the 2D plane. IF-Net texture\cite{chibane_implicit_2021} uses this representation to complete missing colour information in geometrically complete objects. The network is trained to leverage the geometric point's features and adjacent colours to generate a colour-function space. The space can than be sampled at any given point. Other approaches use cascaded 3D convolutional network architectures, which learn to reconstruct corresponding colour information from noisy and imperfect RGB-D maps in a progressive and coarse-to-fine manner \cite{liu_high-quality_2021}. This allows larger missing regions to be reconstructed better.

\begin{figure*}[!h]
    \centering
    \includegraphics[width=\textwidth]{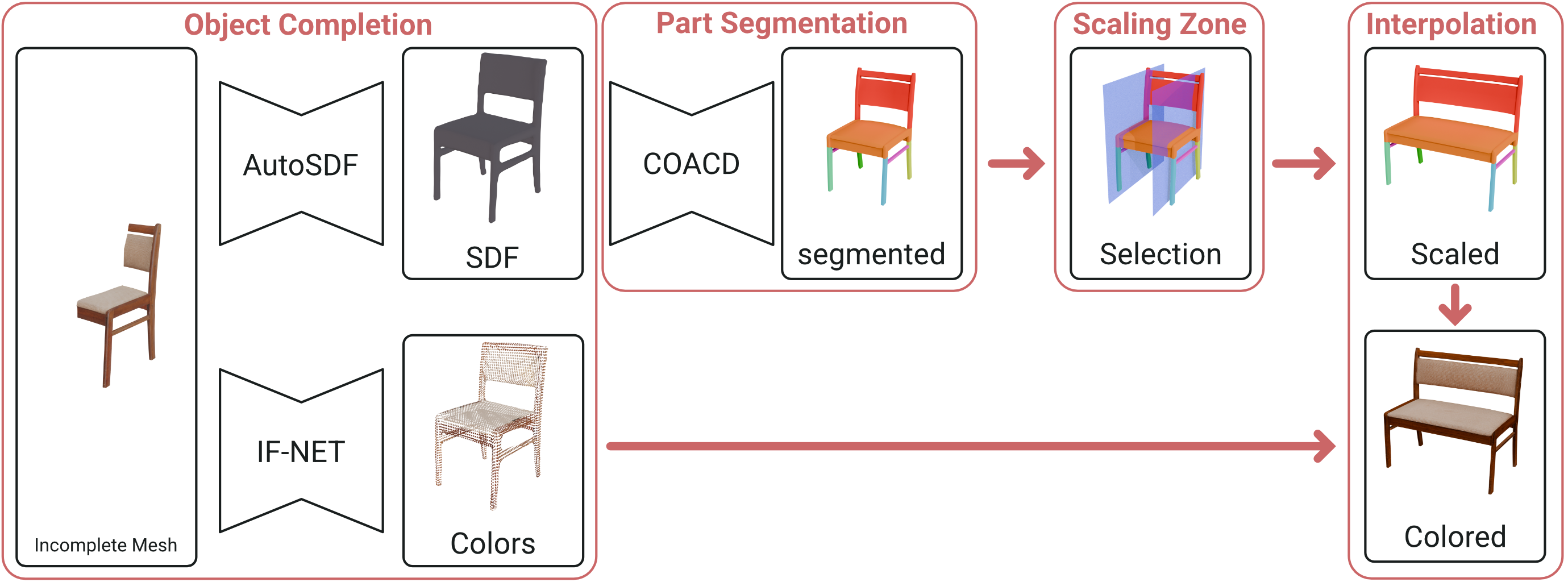}
    \caption{Overview of the proposed pipeline, starting with the object completion (left), going into the part segmentation (centre-left), followed scaling zone definition (centre-right) to result in a scaled and coloured object (right).}
    \label{fig:methodology}
\end{figure*}

\subsection{Part Segmentation}
Object part segmentation enables an object to be divided into smaller parts, either by instance, semantics or both.
CSN \cite{loizou_cross-shape_2023} uses a cross-shape attention mechanism to enable interactions between a shape’s point-wise features and those of other shapes, improving the accuracy and consistency of the shape segmentation.
Mid-Net \cite{wang_unsupervised_2020} uses an unsupervised method for learning a generic and efficient shape encoding network for different shape analysis tasks. The key idea of the method is to jointly encode and learn shape and point features from un-labeled 3D point clouds.
FG-Net \cite{liu_fg-net_2020} is a highly efficient model for large-scale point clouds understanding without voxelizations. it employs a deep convolutional neural network leveraging correlated feature mining and deformable convolution based geometric-aware modelling, in which the local feature relationships and geometric patterns can be fully exploited. These models all aim to segment the models by their semantic label, something which is difficult to generalize for generic objects in a wide variety of scenes.

Approximate convex decomposition (ACD) has become a standard strategy for breaking complex 3D meshes into sets of nearly convex parts, enabling efficient collision detection, physical simulation, and shape analysis. Classical methods such as HACD \cite{mamou_simple_2009} rely on hierarchical clustering with concavity-driven merge heuristics, offering robustness but often producing redundant parts and overlaps. More recently, learning-based methods such as CvxNet \cite{deng_cvxnet_2020} represent shapes as unions of learned convex primitives, achieving compact decompositions but with limited generalization outside the training distribution. To address these challenges, Wei et al. introduced CoACD \cite{wei_approximate_2022}, a geometry-driven algorithm that directly cuts triangle meshes with planes, employs a collision-aware concavity metric sensitive to interior geometry, and explores cut sequences through tree search rather than greedy splitting. This yields intersection-free convex parts with fewer components and higher collision fidelity compared to prior baselines. While these convex parts do not necessarily represent each individual semantic component of an object, the granularity and generalisation of the method ensures the objects are well separated.  

\subsection{Surface-Based Deformation}
Techniques such as 9-slicing \cite{w3_css-backgrounds-3border-image-slice_2024} enable specific zones of images to be scaled while maintaining proportionality in other regions. This method has been extended to 3D environments, such as 27-slicing \cite{deftly_27_2021}, to scale 3D objects without distorting critical regions. However, these methods rely on predefined zones and don't have a scaling constraint, which limits the use of these methods to very regular and basic shapes. KeypointDeformer \cite{jakab_keypointdeformer_2021} tries to combat this by using automatic keypoint detection to guide mesh cage deformation, enabling more natural object transformations. While effective smaller deformations, this method starts to show its limits on large deformations.

Other approaches, like the procedural model generation method described in \cite{getto_automatic_2020}, use automated detection of object components by utilising object skeletons for limiting the deformation regions. Improving its results for more complex objects. These methods all have the same main weakness, that when the parts undergo a large deformation, the selected parts get scales to match the size without any constraint of part size consistency. Other procedural methods, such as Proc-GS \cite{li_proc-gs_2024}, divide 3D models of buildings into components for dynamic recombination. This enables large deformations by introducing modular parts that can be repeated indefinitely to match the desired size. These methods however rely on a clearly defined library of selected parts to build new geometry.

\subsection{SDF-Based Deformation}
SDFs have become a popular representation for 3D object generation due to their standardised size which is ideal for machine learning in-and-outputs. DIF-Net \cite{deng_deformed_2021} represents 3D shapes using a shared template implicit field, deformation fields, and correction fields, enabling non-destructive shape manipulation. Similarly, SALAD \cite{koo_salad_2023} uses a part-level latent diffusion framework for generating and editing 3D shapes, while DualSDF \cite{hao_dualsdf_2020} introduces a two-level SDF representation for semantic shape manipulation. HybridSDF \cite{vasu_hybridsdf_2022} further combines implicit shapes with primitives, allowing a balance between flexibility and structural coherence.

Recent advancements focus on editing SDFs at part and sub-part levels. NVIDIA's XCube \cite{ren_xcube_2023} employs hierarchical voxel latent diffusion models for large-scale 3D generative modeling, enabling low level voxel editing for fine object generation. SPAGHETTI \cite{hertz_spaghetti_2022} enables part-level affine transformations of implicit shapes, such as rotation and translation, while ensuring smooth transitions. However, free scaling at a sub-part level remains a challenge. SENS \cite{binninger_sens_2024} extends SPAGHETTI \cite{hertz_spaghetti_2022} by enabling sketch-based SDF editing but remains limited in its ability to manipulate large-scale deformations without predefined inputs.

\subsection{Texture Deformation}

Traditional mesh textures rely on either UV maps paired with 2D images or per-vertex colour information. Both methods implicitly bind texture appearance to the geometry of the mesh: when the mesh is deformed or scaled, the UV coordinates deform with it, while vertex colours simply interpolate across the new surface. However, image-based textures degrade under large deformations due to stretching or loss of resolution, and vertex colours lack the detail needed to represent high-frequency appearance.

Texture Fields \cite{oechsle_texture_2019} address these limitations by representing texture as a continuous function defined in 3D space, rather than on the surface of the mesh. This functional representation is conceptually similar to SDFs, as both are spatially defined over the object's volume. Because texture values are queried directly in 3D space, Texture Fields can naturally adapt to meshes of different shapes or scales, enabling consistent texture remapping under deformation without loss of detail.

Our work builds on these foundations by combining part-aware SDF deformation with selective scaling capabilities, enabling proportional scaling of arbitrary objects without relying on predefined part databases. This approach bridges the gap between procedural methods and SDF-based editing, providing a robust solution for dynamic object manipulation in indoor scenes.

%=====================METHODOLOGY=====================%

\section{Methodology}
\label{sec:methodology}

The proposed method enables dynamic manipulation of completed objects through a series of steps, including object completion, part detection, scaling zone definition, and the interpolation of SDFs, color, and part indices as illustrated in Fig. \ref{fig:methodology}.

\subsection{Preprocessing: Generating CSDFs}

\begin{figure}[!h]
    \centering
    \includegraphics[width=\columnwidth]{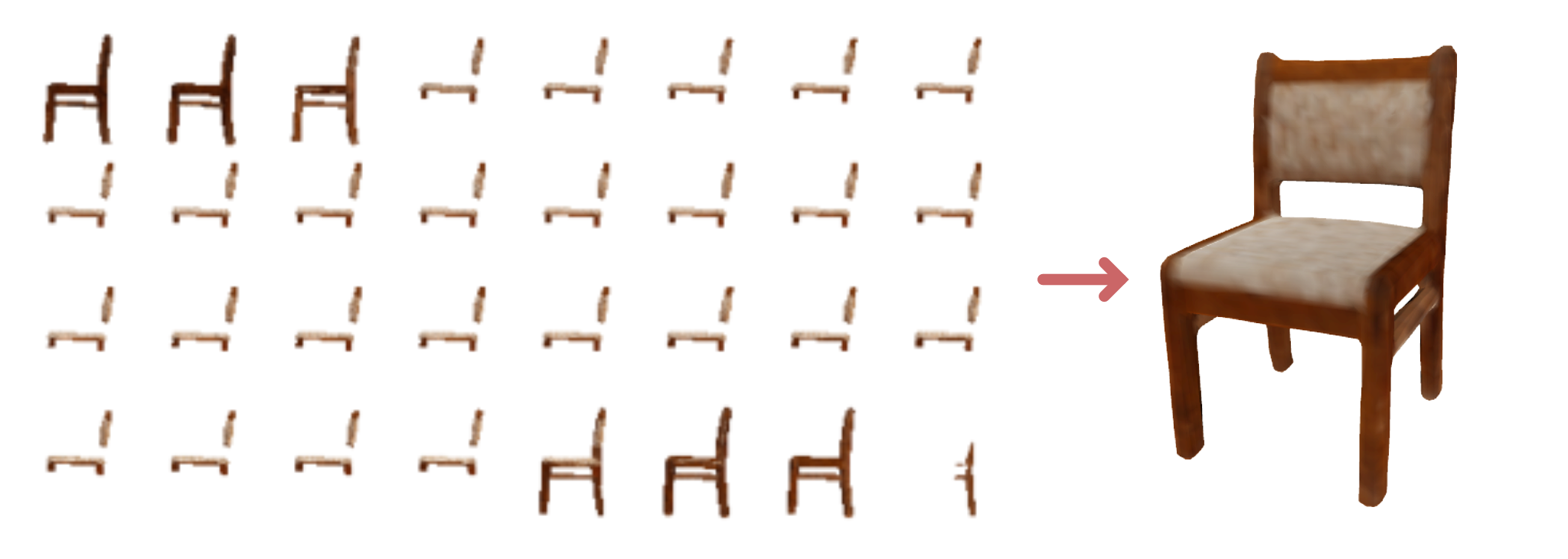}
    \caption{The CSDF stored as a 3D texture, sliced in an 8x8 grid (left) and the rendered object (right)}
    \label{fig:methodology-3Dtexture}
\end{figure}

As a preprocessing step, we complete partially scanned objects from indoor scenes using state-of-the-art methods. Autosdf \cite{mittal_autosdf_2022} is employed to recover full object geometry, while TextureFields \cite{oechsle_texture_2019} is used to infer and complete texture information. The resulting signed distance function (SDF) is discretized into a $128 \times 128 \times 128$ voxel grid, and the corresponding color information is mapped onto the same grid. By combining the geometric SDF with the completed color field, we obtain a colored signed distance function (CSDF) representation. This unified voxel-based CSDF representation serves as the foundation for subsequent processing and transformations, capturing both shape and appearance in a consistent format. The CSDF is then stored as a 3D texture as seen in Figure \ref{fig:methodology-3Dtexture}, where each pixel's rgb value represents the colour information and the alpha channel is reserved for the distance values. 

\subsection{Part Segmentation}

\begin{figure}[!h]
    \centering
    \includegraphics[width=\columnwidth]{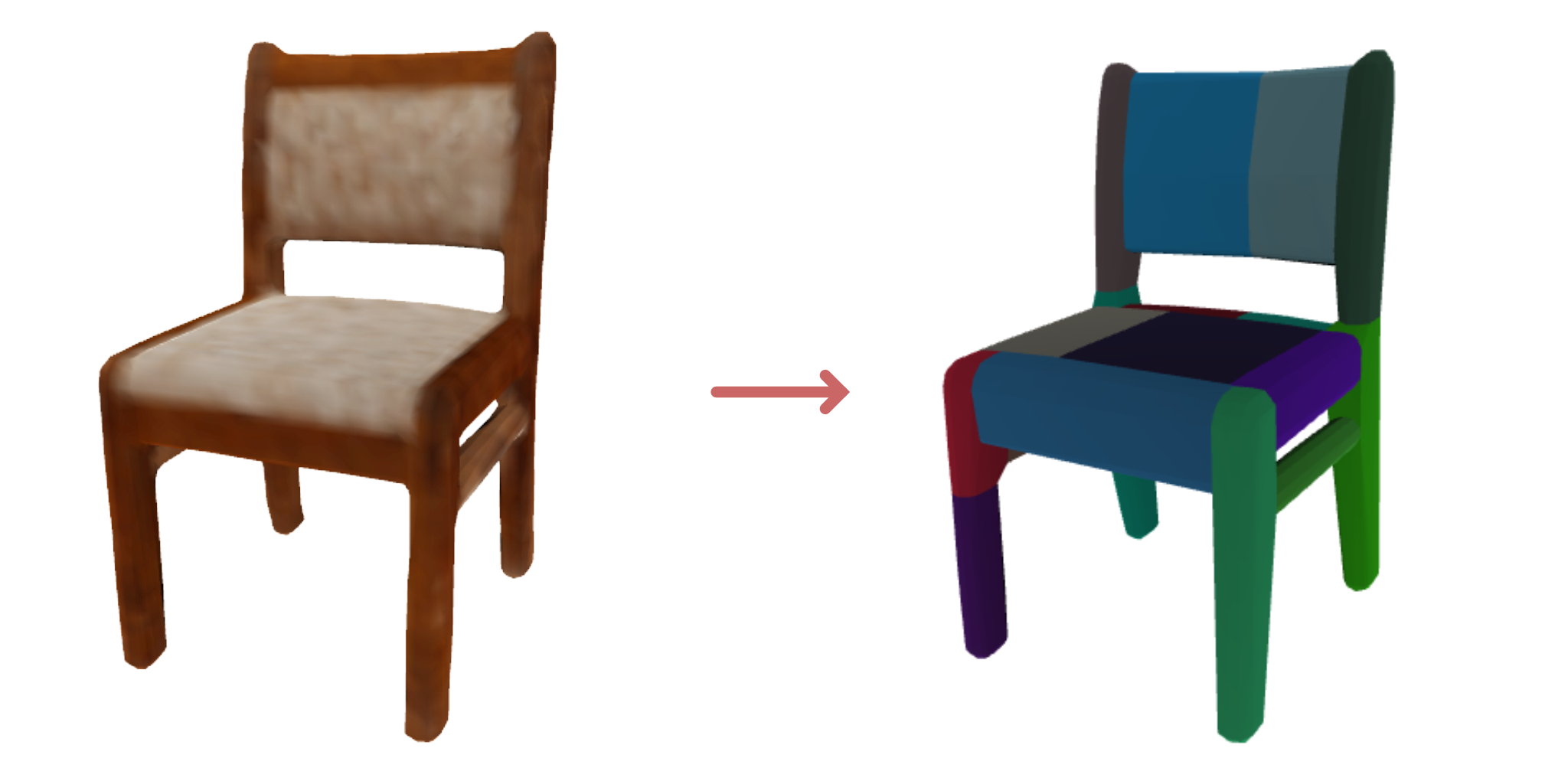}
    \caption{The CSDF rendered as a 3D texture(left) and the convex decomposition of the object (right).}
    \label{fig:methodology-decomposition}
\end{figure}

After the object is reconstructed, it is segmented into distinct parts using Approximate Convex Decomposition \cite{wei_approximate_2022}. A watertight mesh is generated from the SDF using Marching Cubes. The mesh is then separated into different almost-convex parts using cutting planes. By employing a multi-step tree search to determine the cutting planes, the number of unnecessary cuts is greatly reduced. To reduce the parts even further, all pairs of adjacent components are traversed and checked for concavity. If the combined pair is still convex, it is merged. Figure \ref{fig:methodology-decomposition} shows an example of the convex decomposition.

This geometry-based approach allows a more general segmentation not linked to any object class. The convex decomposition also ensures there are no internal intersections and that the interpolation in the next steps can be performed smoothly. After the decomposition, the Segment indices are remapped to the CSDF using nearest-neighbour sampling.
At each voxel coordinate, the object is now represented by three key values: the distance to the nearest surface (SDF), the colour (rgb) and part index (i) at nearest surface.

\subsection{Scaling Zone Definition}

With the object segmented, the next step is to define the scaling zone. This involves selecting specific regions of the object that should be scaled or moved. The selection is determined by defining two parallel planes along a chosen axis: a starting plane and an ending plane. The user is then able to move the ending plane to a new location along the axis. This defines the transformation size.

This planar definition provides a clear and flexible framework for selecting and transforming specific zones of the object without affecting unrelated parts. Furthermore the hands on controls provides very user-friendly controls in a number of different use cases like real-time remodelling or mixed reality applications.

\subsection{SDF, Color, and Part Index Interpolation}

Once the scaling zone is defined, the selection can be scaled. The ending plane is repositioned to its new location, and parts of the object located beyond the ending plane are translated accordingly. Parts located either partially or completely within the scaling zone are isolated for scaling, while those before the starting plane remain unaltered. For the isolated region, a new empty voxel array is created to accommodate the adjusted size of the scaling zone. To perform transformations within the scaling zone, we interpolate the Signed Distance Function (SDF), color values, and part indices. The interpolation techniques vary depending on the type of data being interpolated.

\subsubsection{Linear Interpolation of SDF}

The SDF values are smooth floating-point values, allowing direct linear interpolation. For two SDF values, \(\text{SDF}_1\) and \(\text{SDF}_2\), at positions \(x_1\) and \(x_2\), the interpolated SDF value at any position \(x\) within the range \([x_1, x_2]\) is given by:
\[
\text{SDF}(x) = \text{SDF}_1 + \frac{x - x_1}{x_2 - x_1} (\text{SDF}_2 - \text{SDF}_1)
\]
Here, \(\frac{x - x_1}{x_2 - x_1}\) represents the interpolation factor, which linearly weights the contributions of \(\text{SDF}_1\) and \(\text{SDF}_2\) based on the relative position of \(x\).

After the SDF values are interpolated, the whole SDF is reevaluated and checked for discontinuities. This is performed by recalculating the closest distance to the new surface from each voxel which original distance was greater than the distance to the original scaling zone boundaries. This limitation ensures only distances that could have been changed are recalculated.

\subsubsection{Linear Interpolation of Colour}

The color values are represented in RGB format, where each channel lies within the range \([0, 1]\). Given two colours \(\mathbf{C}_1 = (R_1, G_1, B_1)\) and \(\mathbf{C}_2 = (R_2, G_2, B_2)\), the interpolated colour \(\mathbf{C}(x)\) at position \(x\) is computed as:
\[
\mathbf{C}(x) = \mathbf{C}_1 + \frac{x - x_1}{x_2 - x_1} (\mathbf{C}_2 - \mathbf{C}_1)
\]

Since the colors are discretised in a voxel array, only the voxels near the surface need to be represented. To save on processing costs, all voxels outside the objects boundaries are ignored and left blank. This also saves storage space on the 3D textures.

\subsubsection{Handling Part Index Interpolation}

The part index values are discrete and cannot be interpolated directly. Instead, transitions between part indices are determined by identifying the boundaries where the index changes. Given part indices \(p_1\) and \(p_2\) for regions defined by \([x_1, x_b]\) and \([x_b, x_2]\), the The boundary \(x_b\) is defined as:
\[
x_b = \text{argmin}_x \left( |\text{SDF}(x)| \; \text{where } p(x) \neq p(x+\delta x) \right)
\]
Here, \(\delta x\) is a small step size used to detect transitions between part indices.

\subsection{SDF, Color, and Part Index Repetition}

\begin{figure}[!h]
    \centering
    \includegraphics[width=\columnwidth]{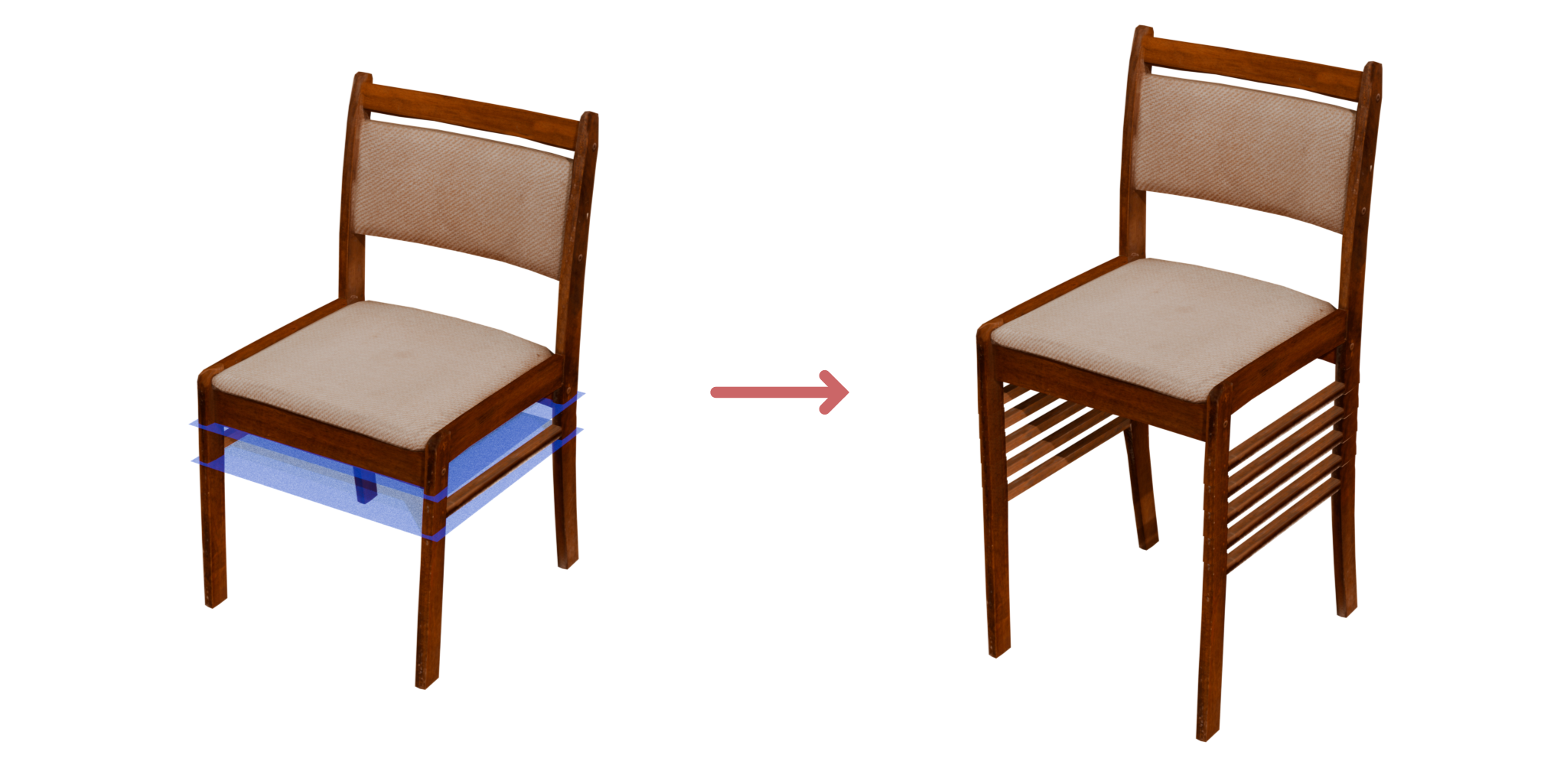}
    \caption{The object with a small repeatable part selected(left) and the resulting repeated parts in the scaled object (right).}
    \label{fig:methodology-repeating}
\end{figure}

For larger deformation a regular stretching of the section will not return the best results. For cases where a repeatable part is scaled, instead of stretching the one part, the selection is repeated a specific number of times based on the total distance as seen in Fig.\ref{fig:methodology-repeating}. To fit the new length precisely, the repeated sections are scaled up or down, with linear interpolation applied as seen in the previous part to maintain continuity. This mode is particularly effective for objects with repetitive patterns, as it prevents excessive distortion by introducing repeated elements instead of stretching the existing geometry.

To ensure the top and bottom parts of the geometry align, the last layers of the interpolated values are smoothed to match the first layers of the next part. This is required for all except the last part, where the end connects back to its original neighbouring part.

\subsection{Saving the Scaled CSDF}
After all these steps, the scaled CSDF is saved again as a 3D texture. For more efficient storing and saving, the CSDF is resampled to match one of the predefined output dimensions.

%=====================EXPERIMENTS=====================%

\section{Experiments}  
\label{sec:experiments}

\begin{figure}[!h]
    \centering
    \includegraphics[width=\columnwidth]{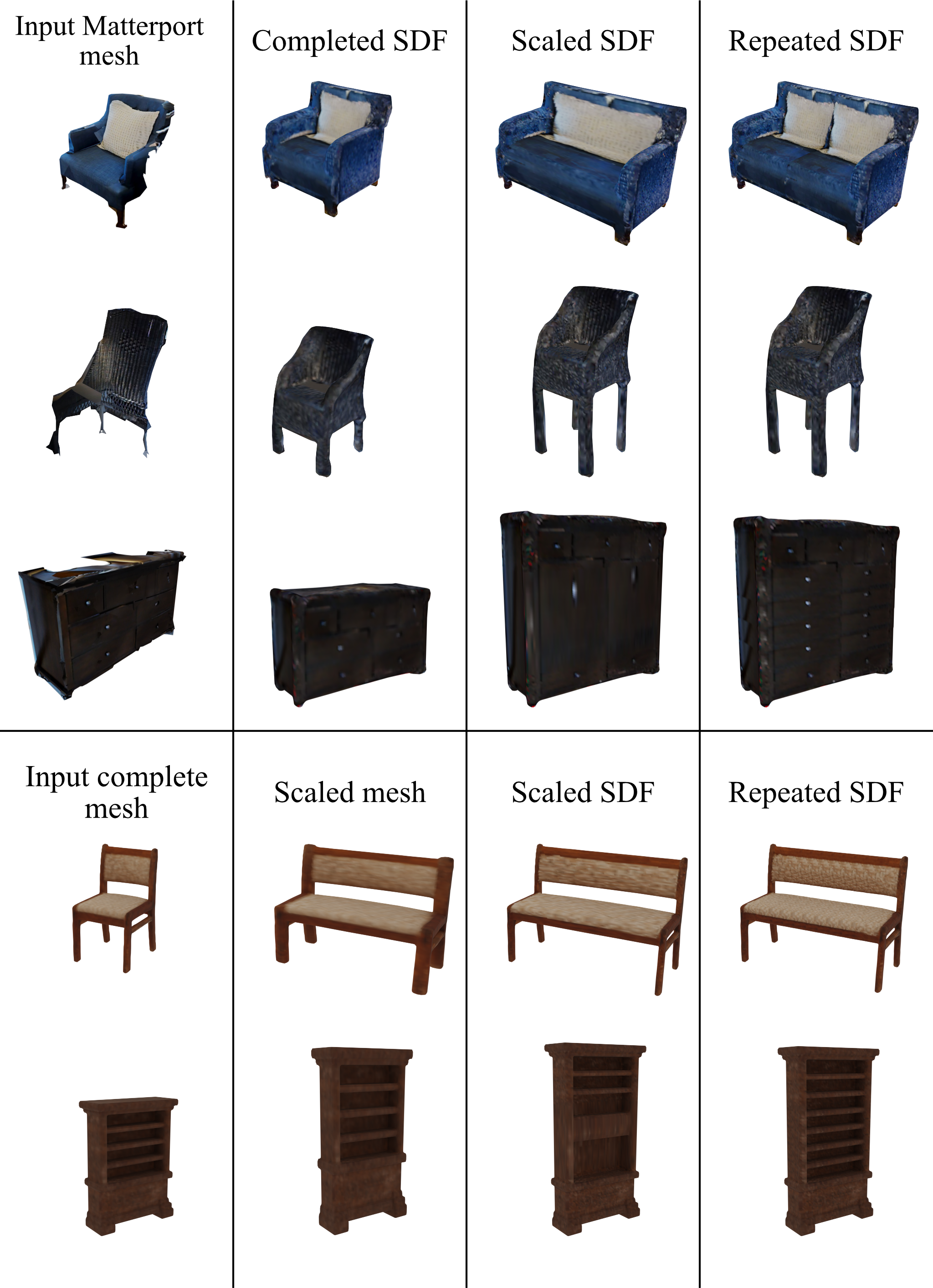}
    \caption{The results of the object scaling pipeline. top: The matterport objects, bottom: the Sketchfab objects.}
    \label{fig:experiments_results}
\end{figure}

To evaluate the effectiveness of our proposed method, we conducted experiments using 3D objects from two main sources. The first dataset was the Matterport 3D Indoor Dataset \cite{chang_matterport3d_2017}, where we extracted objects by isolating them from their original scenes. These objects were then completed both geometrically and texturally using AutoSDF \cite{mittal_autosdf_2022} and IF-Net\cite{chibane_implicit_2021} to create complete objects (Fig.~\ref{fig:experiments_results}, Col. 2, top). Additionally, we curated a set of high-quality, complete 3D scans from Sketchfab for more refined and optimized objects (Fig.~\ref{fig:experiments_results}, Col. 1, bottom).

\begin{figure}[!h]
    \centering
    \includegraphics[width=1\linewidth]{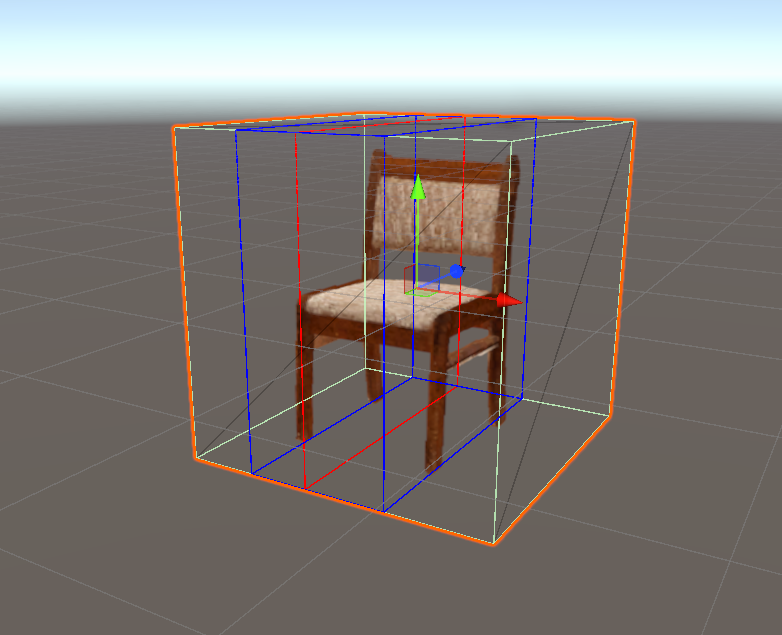}
    \caption{A screenshot of the Unity interface with the three planes defining the start, end, and destination planes in blue, red and blue respectively.}
    \label{fig:experiments_unity}
\end{figure}

The completed CSDFs, stored as 3D textures, were visualised in the Unity game engine using a raymarching Shader \cite{zhou_real-time_2008}. To enable a user-friendly scaling experience, the selection planes were also visualised and can be manipulated using a user interface as seen in Fig.~\ref{fig:experiments_unity}. This ensured realtime feedback of the transformations.

We compared our two selective scaling methods against global objects scaling as seen in Fig.~\ref{fig:experiments_results}. The first method, global object scaling, uniformly stretched the entire object to the desired size. This approach often resulted in unnatural proportions, such as widened chair legs or increased distances between shelves as seen in Fig.~\ref{fig:experiments_results}, Col. 2, bottom. 

The second method, selective object scaling, adjusted only specific parts of the object based on the predefined scaling zones. This approach performed well for regular shapes, like chair legs and seats, but failed with complex geometries constrained by multiple directions, such as shelves or cushions as seen in Fig.~\ref{fig:experiments_results}, Col. 3. 

Finally, selective object tiling preserved proportions by repeating specific components of the object. This method excelled for repeating patterns, such as bookshelf shelves or sofa cushions, maintaining structural integrity and realistic appearance as seen in Fig.~\ref{fig:experiments_results}, Col. 4.  

\section{Discussion}

Our results highlight several important observations. First, selective scaling offers a clear advantage over global object scaling by preserving proportions in regions outside the scaling zone. This is particularly evident for objects whose functional or aesthetic properties depend on localized geometry, such as chair legs or table surfaces. However, selective scaling alone introduces noticeable distortions in objects with complex interdependent structures, where multiple constraints interact across different axes. In such cases, deformation accumulates and results in implausible geometry.

The selective tiling approach addresses this limitation by repeating geometric and textural patterns instead of stretching them. This strategy performs especially well for objects composed of modular elements—such as shelves, cabinets, or cushioned furniture—where repetition is structurally meaningful. Tiling maintains appearance fidelity, reduces stretching artifacts, and preserves the consistency of repeated subcomponents. Nonetheless, tiling is less suitable for objects lacking clear repeatable regions, or for those with intricate or highly organic shapes where repetition becomes visually inconsistent.

Another key insight concerns the CSDF representation itself. Storing both SDF and color fields in a unified voxel grid enables smooth joint interpolation of geometry and appearance but introduces a trade-off between memory footprint and representational fidelity. While the chosen $128^3$ resolution provides a good balance, finer details may still be lost during interpolation or resampling. Moreover, although convex decomposition provides class-agnostic segmentation suitable for general manipulation, its reliance on geometric convexity can produce segmentations that differ from semantic parts in some cases.

Overall, the experiments validate the practicality of our approach for interactive editing scenarios, such as real-time object customization or mixed-reality applications. The ability to define scaling zones using simple planar controls supports intuitive manipulation, and the integration with Unity allows immediate visual feedback through raymarched rendering. Future work could improve segmentation robustness, investigate adaptive voxel resolutions, or integrate learned deformation priors to further enhance manipulation quality on complex objects.

%=====================CONCLUSIONS=====================%
\section{Conclusion}
\label{sec:conclusion}

In this paper, we presented a novel method for dynamic object manipulation through context-aware scaling. By integrating object completion, part segmentation, scaling zone definitions, and advanced interpolation techniques, our method achieves proportional and visually coherent object scaling. The use of SDF's for surface definitions and Texturefields for color representations, The scaling can be done intuitively and consistently. Additionally, the inclusion of repeating patterns ensures structural integrity during large deformations, addressing a critical gap in existing scaling methodologies.  

The integration of our CSDF pipeline into a real-time Unity interface further highlights its potential for interactive editing, rapid prototyping, and mixed-reality applications. While limitations remain for objects with highly complex or non-repetitive structures, the results show that CSDF-driven manipulation provides a powerful, unified representation for consistent geometry and texture deformation.

Future work will explore different strategies to both enable more organic shapes to be scaled realistically and allow for more organic transformations of rigid shapes.

{
	\begin{spacing}{1.17}
		\normalsize
		\bibliography{Dynamic_Object_Completion} % Include your own bibliography (*.bib), style is given in isprs.cls

@article{liu_fg-net_2020,
	title = {{FG}-{Net}: {Fast} {Large}-{Scale} {LiDAR} {Point} {Clouds} {Understanding} {Network} {Leveraging} {Correlated} {Feature} {Mining} and {Geometric}-{Aware} {Modelling}},
	url = {http://arxiv.org/abs/2012.09439},
	author = {Liu, Kangcheng and Gao, Zhi and Lin, Feng and Chen, Ben M.},
	month = dec,
	year = {2020},
}

@article{wang_unsupervised_2020,
	title = {Unsupervised {3D} {Learning} for {Shape} {Analysis} via {Multiresolution} {Instance} {Discrimination}},
	url = {http://arxiv.org/abs/2008.01068},
	author = {Wang, Peng-Shuai and Yang, Yu-Qi and Zou, Qian-Fang and Wu, Zhirong and Liu, Yang and Tong, Xin},
	month = aug,
	year = {2020},
}

@article{loizou_cross-shape_2023,
	title = {Cross-{Shape} {Attention} for {Part} {Segmentation} of {3D} {Point} {Clouds}},
	volume = {42},
	issn = {14678659},
	doi = {10.1111/cgf.14909},
	number = {5},
	journal = {Computer Graphics Forum},
	author = {Loizou, Marios and Garg, Siddhant and Petrov, Dmitry and Averkiou, Melinos and Kalogerakis, Evangelos},
	month = aug,
	year = {2023},
	note = {Publisher: John Wiley and Sons Inc},
}

@misc{w3_css-backgrounds-3border-image-slice_2024,
	title = {css-backgrounds-3/\#border-image-slice},
	url = {https://www.w3.org/TR/css-backgrounds-3/#border-image-slice},
	author = {{W3}},
	year = {2024},
}

@article{getto_automatic_2020,
	title = {Automatic procedural model generation for {3D} object variation},
	volume = {36},
	issn = {01782789},
	doi = {10.1007/s00371-018-1589-4},
	number = {1},
	journal = {Visual Computer},
	author = {Getto, Roman and Kuijper, Arjan and Fellner, Dieter W.},
	month = jan,
	year = {2020},
	note = {Publisher: Springer},
	pages = {53--70},
}

@article{li_proc-gs_2024,
	title = {Proc-{GS}: {Procedural} {Building} {Generation} for {City} {Assembly} with {3D} {Gaussians}},
	url = {http://arxiv.org/abs/2412.07660},
	author = {Li, Yixuan and Ran, Xingjian and Xu, Linning and Lu, Tao and Yu, Mulin and Wang, Zhenzhi and Xiangli, Yuanbo and Lin, Dahua and Dai, Bo},
	month = dec,
	year = {2024},
}

@article{binninger_sens_2024,
	title = {{SENS}: {Part}-{Aware} {Sketch}-based {Implicit} {Neural} {Shape} {Modeling}},
	volume = {43},
	issn = {14678659},
	doi = {10.1111/cgf.15015},
	number = {2},
	journal = {Computer Graphics Forum},
	author = {Binninger, Alexandre and Hertz, Amir and Sorkine-Hornung, Olga and Cohen-Or, Daniel and Giryes, Raja},
	month = may,
	year = {2024},
	note = {Publisher: John Wiley and Sons Inc},
}

@article{jakab_keypointdeformer_2021,
	title = {{KeypointDeformer}: {Unsupervised} {3D} {Keypoint} {Discovery} for {Shape} {Control}},
	url = {http://arxiv.org/abs/2104.11224},
	author = {Jakab, Tomas and Tucker, Richard and Makadia, Ameesh and Wu, Jiajun and Snavely, Noah and Kanazawa, Angjoo},
	month = apr,
	year = {2021},
}

@inproceedings{vasu_hybridsdf_2022,
	title = {{HybridSDF}: {Combining} {Deep} {Implicit} {Shapes} and {Geometric} {Primitives} for {3D} {Shape} {Representation} and {Manipulation}},
	isbn = {978-1-6654-5670-8},
	doi = {10.1109/3DV57658.2022.00072},
	booktitle = {Proceedings - 2022 {International} {Conference} on {3D} {Vision}, {3DV} 2022},
	publisher = {Institute of Electrical and Electronics Engineers Inc.},
	author = {Vasu, Subeesh and Talabot, Nicolas and Lukoianov, Artem and Baque, Pierre and Donier, Jonathan and Fua, Pascal},
	year = {2022},
	pages = {617--626},
}

@article{hertz_spaghetti_2022,
	title = {{SPAGHETTI}: {Editing} {Implicit} {Shapes} {Through} {Part} {Aware} {Generation}},
	volume = {41},
	issn = {15577368},
	doi = {10.1145/3528223.3530084},
	number = {4},
	journal = {ACM Transactions on Graphics},
	author = {Hertz, Amir and Perel, Or and Giryes, Raja and Sorkine-Hornung, Olga and Cohen-Or, Daniel},
	month = jul,
	year = {2022},
	note = {Publisher: Association for Computing Machinery},
}

@inproceedings{koo_salad_2023,
	title = {{SALAD}: {Part}-{Level} {Latent} {Diffusion} for {3D} {Shape} {Generation} and {Manipulation}},
	url = {https://salad3d.github.io.},
	author = {Koo, Juil and Yoo, Seungwoo and Nguyen, Minh Hieu},
	year = {2023},
}

@inproceedings{deng_deformed_2021,
	title = {Deformed {Implicit} {Field}: {Modeling} {3D} {Shapes} with {Learned} {Dense} {Correspondence} {Embedded} {Shape} {Texture} {Transfer} {Shape} {Editing}},
	url = {https://github.com/microsoft/DIF-Net.},
	author = {Deng, Yu and Yang, Jiaolong and Tong, Xin},
	year = {2021},
}

@inproceedings{hao_dualsdf_2020,
	title = {{DualSDF}: {Semantic} {Shape} {Manipulation} using a {Two}-{Level} {Representation}},
	url = {https://github.com/zekunhao1995/},
	author = {Hao, Zekun and Averbuch-Elor, Hadar and Snavely, Noah and Belongie, Serge},
	year = {2020},
}

@article{ren_xcube_2023,
	title = {{XCube} ({\textbackslash}mathcal\{{X}\}{\textasciicircum}3): {Large}-{Scale} {3D} {Generative} {Modeling} using {Sparse} {Voxel} {Hierarchies}},
	url = {http://arxiv.org/abs/2312.03806},
	author = {Ren, Xuanchi and Huang, Jiahui and Zeng, Xiaohui and Museth, Ken and Fidler, Sanja and Williams, Francis},
	month = dec,
	year = {2023},
}

@article{chang_matterport3d_2017,
	title = {{Matterport3D}: {Learning} from {RGB}-{D} {Data} in {Indoor} {Environments}},
	journal = {International Conference on 3D Vision (3DV)},
	author = {Chang, Angel and Dai, Angela and Funkhouser, Thomas and Halber, Maciej and Niessner, Matthias and Savva, Manolis and Song, Shuran and Zeng, Andy and Zhang, Yinda},
	year = {2017},
}

@misc{chibane_implicit_2021,
	title = {Implicit {Feature} {Networks} for {Texture} {Completion} from {Partial} {3D} {Data}},
	url = {https://cvi2.uni.lu/sharp2020/},
	author = {Chibane, Julian and Pons-Moll, Gerard},
	year = {2021},
}

@article{oechsle_texture_2019,
	title = {Texture {Fields}: {Learning} {Texture} {Representations} in {Function} {Space}},
	url = {http://arxiv.org/abs/1905.07259},
	author = {Oechsle, Michael and Mescheder, Lars and Niemeyer, Michael and Strauss, Thilo and Geiger, Andreas},
	month = may,
	year = {2019},
}

@article{mittal_autosdf_2022,
	title = {{AutoSDF}: {Shape} {Priors} for {3D} {Completion}, {Reconstruction} and {Generation}},
	url = {http://arxiv.org/abs/2203.09516},
	author = {Mittal, Paritosh and Cheng, Yen-Chi and Singh, Maneesh and Tulsiani, Shubham},
	month = mar,
	year = {2022},
}

@article{wei_approximate_2022,
	title = {Approximate {Convex} {Decomposition} for {3D} {Meshes} with {Collision}-{Aware} {Concavity} and {Tree} {Search}},
	volume = {41},
	issn = {0730-0301, 1557-7368},
	url = {http://arxiv.org/abs/2205.02961},
	doi = {10.1145/3528223.3530103},
	language = {en},
	number = {4},
	urldate = {2025-09-30},
	journal = {ACM Transactions on Graphics},
	author = {Wei, Xinyue and Liu, Minghua and Ling, Zhan and Su, Hao},
	month = jul,
	year = {2022},
	note = {arXiv:2205.02961 [cs]},
	pages = {1--18},
}

@misc{deftly_27_2021,
	title = {27 {Slicer}},
	url = {https://slicer.deftly.games/},
	urldate = {2025-08-24},
	publisher = {Deftly Games},
	author = {{Deftly}},
	year = {2021},
}

@inproceedings{mamou_simple_2009,
	address = {Cairo, Egypt},
	title = {A simple and efficient approach for {3D} mesh approximate convex decomposition},
	isbn = {978-1-4244-5653-6},
	url = {http://ieeexplore.ieee.org/document/5414068/},
	doi = {10.1109/ICIP.2009.5414068},
	language = {en},
	urldate = {2025-09-30},
	booktitle = {2009 16th {IEEE} {International} {Conference} on {Image} {Processing} ({ICIP})},
	publisher = {IEEE},
	author = {Mamou, Khaled and Ghorbel, Faouzi},
	month = nov,
	year = {2009},
	pages = {3501--3504},
}

@inproceedings{deng_cvxnet_2020,
	address = {Seattle, WA, USA},
	title = {{CvxNet}: {Learnable} {Convex} {Decomposition}},
	copyright = {https://ieeexplore.ieee.org/Xplorehelp/downloads/license-information/IEEE.html},
	isbn = {978-1-7281-7168-5},
	shorttitle = {{CvxNet}},
	url = {https://ieeexplore.ieee.org/document/9157370/},
	doi = {10.1109/CVPR42600.2020.00011},
	language = {en},
	urldate = {2025-09-30},
	booktitle = {2020 {IEEE}/{CVF} {Conference} on {Computer} {Vision} and {Pattern} {Recognition} ({CVPR})},
	publisher = {IEEE},
	author = {Deng, Boyang and Genova, Kyle and Yazdani, Soroosh and Bouaziz, Sofien and Hinton, Geoffrey and Tagliasacchi, Andrea},
	month = jun,
	year = {2020},
	pages = {31--41},
}

@inproceedings{vermandere_geometry_2025,
	address = {Porto, Portugal},
	title = {Geometry and {Texture} {Completion} of {Partially} {Scanned} {3D} {Objects} {Through} {Material} {Segmentation}:},
	isbn = {978-989-758-728-3},
	shorttitle = {Geometry and {Texture} {Completion} of {Partially} {Scanned} {3D} {Objects} {Through} {Material} {Segmentation}},
	url = {https://www.scitepress.org/DigitalLibrary/Link.aspx?doi=10.5220/0013120000003912},
	doi = {10.5220/0013120000003912},
	language = {en},
	urldate = {2025-10-27},
	booktitle = {Proceedings of the 20th {International} {Joint} {Conference} on {Computer} {Vision}, {Imaging} and {Computer} {Graphics} {Theory} and {Applications}},
	publisher = {SCITEPRESS - Science and Technology Publications},
	author = {Vermandere, Jelle and Bassier, Maarten and Vergauwen, Maarten},
	year = {2025},
	pages = {193--202},
}

@article{vermandere_guided_2025,
	title = {Guided object completion with interactive voxel editing},
	volume = {X-G-2025},
	copyright = {https://creativecommons.org/licenses/by/4.0/},
	issn = {2194-9050},
	url = {https://isprs-annals.copernicus.org/articles/X-G-2025/901/2025/},
	doi = {10.5194/isprs-annals-X-G-2025-901-2025},
	language = {en},
	urldate = {2025-10-27},
	journal = {ISPRS Annals of the Photogrammetry, Remote Sensing and Spatial Information Sciences},
	author = {Vermandere, Jelle and Bassier, Maarten and Vergauwen, Maarten},
	month = jul,
	year = {2025},
	pages = {901--906},
}

@article{liu_high-quality_2021,
	title = {High-{Quality} {Textured} {3D} {Shape} {Reconstruction} with {Cascaded} {Fully} {Convolutional} {Networks}},
	volume = {27},
	copyright = {https://ieeexplore.ieee.org/Xplorehelp/downloads/license-information/IEEE.html},
	issn = {1077-2626, 1941-0506, 2160-9306},
	url = {https://ieeexplore.ieee.org/document/8812900/},
	doi = {10.1109/TVCG.2019.2937300},
	language = {en},
	number = {1},
	urldate = {2025-10-27},
	journal = {IEEE Transactions on Visualization and Computer Graphics},
	author = {Liu, Zheng-Ning and Cao, Yan-Pei and Kuang, Zheng-Fei and Kobbelt, Leif and Hu, Shi-Min},
	month = jan,
	year = {2021},
	pages = {83--97},
}

@article{zhou_real-time_2008,
	title = {Real-time smoke rendering using compensated ray marching},
	volume = {27},
	issn = {0730-0301, 1557-7368},
	url = {https://dl.acm.org/doi/10.1145/1360612.1360635},
	doi = {10.1145/1360612.1360635},
	language = {en},
	number = {3},
	urldate = {2025-11-17},
	journal = {ACM Transactions on Graphics},
	author = {Zhou, Kun and Ren, Zhong and Lin, Stephen and Bao, Hujun and Guo, Baining and Shum, Heung-Yeung},
	month = aug,
	year = {2008},
	pages = {1--12},
}
	\end{spacing}
}

\end{document}